\documentclass[conference]{IEEEtran}

\usepackage{amsfonts}
\usepackage{algorithmic}
\usepackage{graphicx}
\usepackage{textcomp}
\usepackage{xcolor}
\usepackage{ulem}
\usepackage{url}
\usepackage{subcaption} 
\usepackage{multirow}

\title{Exploring Robustness of Image Recognition Models on Hardware Accelerators}

\author{\IEEEauthorblockN{Nikolaos Louloudakis}
\IEEEauthorblockA{\textit{n.louloudakis@ed.ac.uk} \\
\textit{University of Edinburgh} \\
\textit{United Kingdom\vspace{-3.2\baselineskip}}} \\
\and
\IEEEauthorblockN{Perry Gibson}
\IEEEauthorblockA{\textit{perry.gibson@glasgow.ac.uk} \\
\textit{University of Glasgow} \\
\textit{United Kingdom\vspace{-3.2\baselineskip}}} \\
\and
\IEEEauthorblockN{Jos\'e Cano}
\IEEEauthorblockA{\textit{jose.canoreyes@glasgow.ac.uk} \\
\textit{University of Glasgow} \\
\textit{United Kingdom\vspace{-3.2\baselineskip}}} \\
\and
\IEEEauthorblockN{Ajitha Rajan}
\IEEEauthorblockA{\textit{arajan@ed.ac.uk} \\
\textit{University of Edinburgh} \\
\textit{United Kingdom\vspace{-3.2\baselineskip}}} \\
\thanks{Perry Gibson contributed to this work while at the University of Glasgow.}
}

\IEEEoverridecommandlockouts

\begin{document}

\newcommand{\tool}{\texttt{MutateNN}} 

\maketitle


\begin{abstract}

As the usage of Artificial Intelligence (AI) on resource-intensive and safety-critical tasks increases, a variety of Machine Learning (ML) compilers have been developed, enabling compatibility of Deep Neural Networks (DNNs) with a variety of hardware acceleration devices. However, given that DNNs are widely utilized for challenging and demanding tasks, the behavior of these compilers must be verified.
To this direction, we propose \tool, a tool that utilizes elements of both differential and mutation testing in order to examine the robustness of image recognition models when deployed on hardware accelerators with different capabilities, in the presence of faults in their target device code - introduced either by developers, or problems in their compilation process.
We focus on the image recognition domain by applying mutation testing to $7$ well-established DNN models, introducing $21$ mutations of $6$ different categories. We deployed our mutants on $4$ different hardware acceleration devices of varying capabilities and observed that DNN models presented discrepancies of up to $90.3$\% in mutants related to conditional operators across devices. We also observed that mutations related to layer modification, arithmetic types and input affected severely the overall model performance (up to $99.8$\%) or led to model crashes, in a consistent manner across devices.

\end{abstract}


\section{Introduction}

Enabling optimization and hardware acceleration of Deep Neural Networks (DNNs) is vital to achieve high performance~\cite{gibson_dlas_2024}. Given the complexity of the task, and in order to automate the process of DNN model compilation, optimization and deployment on different hardware acceleration devices, a number of Machine Learning (ML) compilers have been implemented. 
OpenAI Triton~\cite{triton} and Apache TVM~\cite{tvm} allow the compilation and automatic optimization of DNNs, while MLIR~\cite{mlir} provides a low-level framework for developing AI compilers.
However, the compilation process can be error-prone, due to the architectural and structural complexity of DNNs, intertwined with effective hardware acceleration orchestration. A recent study~\cite{humbatova2020taxonomy} has shown that GPU bugs are amongst the main faults in DNNs, indicating that the process of integrating hardware acceleration in DNNs needs to be adequately tested.
In addition, our prior research indicates that model performance can unexpectedly vary across GPU devices~\cite{louloudakis2022assessing, deltann}.

For that purpose, we propose \tool, a tool that utilizes differential and mutation testing in order to examine the robustness of compiled DNNs deployed in different hardware acceleration devices in the presence of errors -  potentially introduced by (1) developers writing custom device code, or (2) faults in AI compilers. 
Inspired by mutations generated in conventional software, \tool\ is able to generate DNN model mutants, execute them, and compare their behavior both against the original model, as well as across a variety of hardware acceleration devices. 
\tool\ leverages the popular Apache TVM~\cite{tvm} compiler. To demonstrate the capabilities of \tool, we selected $7$ widely-known image recognition models, which were all pre-trained with the ImageNet~\cite{deng2009imagenet} dataset. 
Inspired by the literature and by conventional mutation testing techniques, we generated $21$ mutations of $6$ different categories: Layer Node Replacement, Arithmetic Node Replacement, Node Input/Output Modifications, Arithmetic Types Mutations, Kernel Variables \& Stores Mutations, and Conditional Statement Operations Mutations.
Using these mutations, we performed inference on $4$ hardware devices of varying computational capabilities. We observed up to $90.3$\% output label prediction differences across mutants related to conditional operations, as well as an unexpected model correctness degradation related to mutants of arithmetic types.

\section{System Architecture}
\label{system-arch}

\begin{figure*}[!t]
 \centering
 \includegraphics[width=\textwidth]{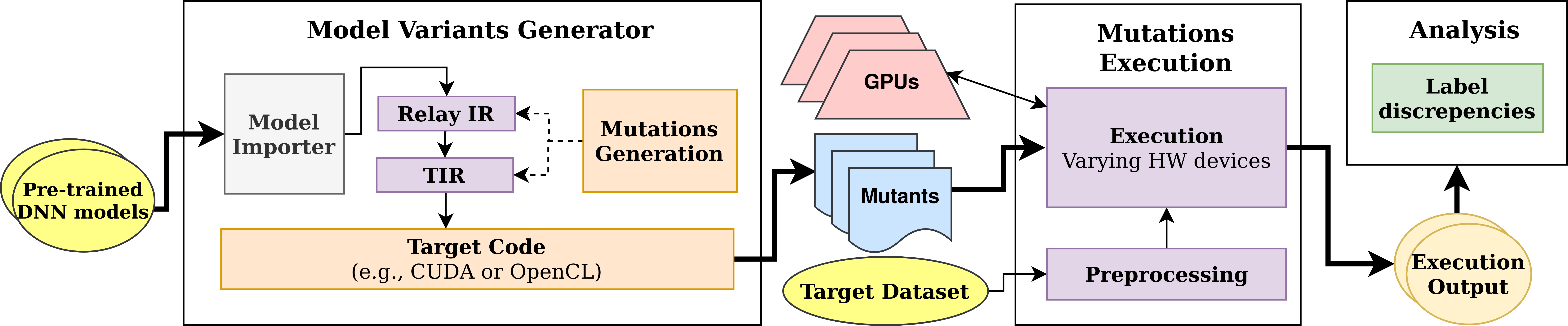}
 \caption{Architecture of \tool: (1) \textit{Model Variants Generator} generates mutations and compiles them to device code; (2) \textit{Mutations Execution} executes the various mutants on images from a target dataset; and (3) \textit{Analysis} compares inference outputs and reports metrics across mutant executions.}
 \label{fig:mutatenn}
 \vspace{-15pt}
\end{figure*}

\tool\ consists of three main components: the \textit{Model Variants Generator Module}, the \textit{Execution Module}, and the \textit{Analysis Module}. 
It utilizes a configuration-based approach, capable of generating, deploying, performing inference and analyzing a batch of model mutations. The system architecture is presented in Figure~\ref{fig:mutatenn}. 
\tool\ is built on top of the \textit{TVM} compiler stack~\cite{tvm}, which allows parameterization of models and deployment on different hardware acceleration devices. \tool\ allows end users to define a configuration of the experiments under test in a JSON file.

\paragraph{Model Variants Generator Module}
To test the behavior of DNNs across different hardware acceleration devices, \tool\ enables the generation of model mutations, focusing on two primary categories: (1) graph-related mutations, implemented at Relay IR (TVM's high-level graph IR); and (2) code-related mutations, implemented at TIR (TVM's low-level IR, one level above device code generation). \tool\ loads a model from its initial format, which can be either a pre-trained model from a well-established library (e.g., Keras~\cite{chollet2015keras}), or a file-based representation, such as ONNX~\cite{onnxsite}. 
\tool\ then compiles the mutants following a user-defined configuration by utilizing TVM, generating an executable targeting a specific device. The output is a DNN model in compressed format that can be loaded via TVM, deployed and executed at a later stage in the process. It is important to highlight that the module will only generate valid models, i.e., that succeed through the compilation process of TVM, otherwise a compilation error is raised and the mutant is not further considered for execution.

\paragraph{Mutations Execution Module}
Once the tool completes the mutant generation process, \tool\ loads and deploys a mutant to a hardware acceleration device for model inference, via Remote Process Communication (RPC).
Initially, \tool\ loads and deploys the model in the hardware acceleration device defined in system configuration. 
Then, it utilizes the evaluation dataset defined in the configuration. In particular, it applies the necessary pre-processing (e.g., normalization) on each dataset image, performs inference against the model mutant, and generates the top-K inference result along with the execution time. Finally, it stores each output to a separate file for further examination by the analysis module.

\paragraph{Analysis Module}
Once the inference operations are complete for the whole experiment set (the original model and its mutants, executed across all devices using the dataset under test), then \tool\ applies analysis to determine discrepancies between mutant and original model executions by utilizing pairwise comparison against the inference outputs generated by different devices for the respective mutant under test. 
In particular, \tool\ loads all files containing the prediction outputs for the dataset, generated for each mutant across devices, and performs pairwise comparison, using a user-selected metric.
Note that \tool\ supports a number of comparison metrics, such as top inference output label comparison, Levenshtein distance and Kendall's Tau. The reporting results are generated in a summary JSON file, enabling further processing.

\paragraph{Configuration}
\tool\ allows its setup to be configured via a JSON file, which contains properties related to: (1) specification of models under test, along with their properties (e.g., input and output dimensions); (2) mutations to be generated, both in graph and kernel code level; (3) devices where the models will be deployed and executed on; and (4) dataset definitions, utilized for differential testing purposes. A sample of the configuration file, along with directions on how it can be utilized, can be found in \tool\ code repository.

\section{Implementation}
\label{implementation}

\tool\ is implemented on top of \texttt{Apache TVM}~\cite{tvm}, a performance-oriented AI compiler stack that enables DNN optimization and deployment in a variety of hardware acceleration devices. \tool\ utilizes TVM Graph (Relay) and Tensor-Level IR (TIR) to generate model mutants with the purpose of testing the AI compiler-generated code that enables hardware acceleration.

\paragraph{Model Variants Generation}
\tool\ is implemented so that users can define and configure their own custom mutants, along with their scope and effect range in the DNN model. 
With regards to Relay IR, \tool\ slightly mutates parts of the model graph. For instance, it introduces nodes (e.g., transpose), affecting model structure and data manipulation. \tool\ applies a best-effort approach to generate valid model graphs, by automatically adjusting the mutated layer arguments, considering their number, type and dimensions in the layer utilized. This is performed by using TVM's type checker API. As about TIR, \tool\ applies modifications related to the kernel code (e.g., conditional statements responsible for thread handling), generated from \texttt{TVM}, in order to enable hardware acceleration for the DNN model.

\begin{figure}[!htp]
 \centering
 \includegraphics[width=\linewidth]{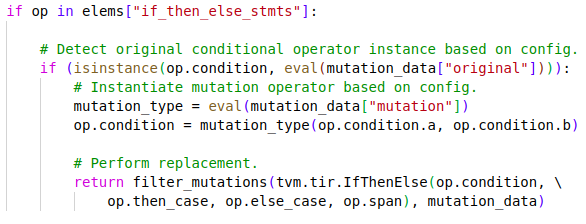}
 \caption{Implementation of the operator mutation generation in TIR pass.} 
 \label{fig:implementation_tir}
 \vspace{-10pt}
\end{figure}

\begin{figure}[!htp]
 \centering
 \includegraphics[width=\linewidth]{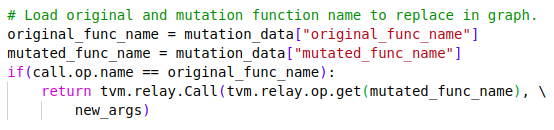}
 \caption{Implementation of the activation function replacement mutation generation in Relay IR pass.}
 \vspace{-15pt}
 
 \label{fig:implementation_relay}
\end{figure}

To briefly demonstrate our approach, we present two samples of code, part of the implementation of TIR pass (Figure~\ref{fig:implementation_tir}), demonstrating the mutation of conditional operators, and Relay IR pass (Figure~\ref{fig:implementation_relay}), demonstrating the replacement of activation functions. 

In the first case, we apply transformations in expression level (i.e., mutating conditional statement operands), while in the second case, we perform a layer replacement in the model graph. Overall, we present the set of mutations supported by \tool\ in the next section.

\paragraph{Mutations Supported}
\tool\ supports two primary types of mutations: \textbf{(1) Graph-Related}, and \textbf{(2) Code-Related}. The mutations are inspired by related work (primarily from Tzer~\cite{TZER} and DeepMutation~\cite{deepmutation}), by an established taxonomy of real faults in DNNs~\cite{humbatova2020taxonomy}, but also from conventional mutation testing applying changes in expressions and types.
In total, we utilize $21$ different  mutations. The aim, is to imitate real-world faults and discover to what extent DNNs are vulnerable to such errors. However, this set of mutations is indicative, and the system can be utilized to generate new mutants with respect to the related configuration. However, our mutations list is far from exhaustive. A considerable portion of operators suggested by the literature~\cite{deepcrime, deepmutation, deepmutationplusplus, TZER} are yet to be implemented in \tool. We selected an indicative subset of popular mutations, as a proof of concept for \tool. 
In addition, in our runs, \tool\ is currently focused in the deployment process, and therefore we disregard any mutations or analysis related to the training process. However, \tool\ supports the definition of additional types of mutations through its configuration, which are subject to future work.

\begin{figure}[!htp]
 \centering
 \includegraphics[width=\linewidth]{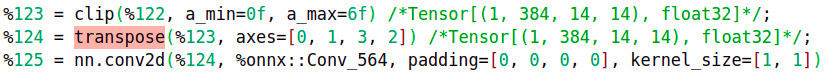}
 \caption{Injected tensor transposition in Relay IR.} 
\label{fig:transpose_example}
\vspace{-15pt}
\end{figure}

For graph-related mutations, we generate mutants based on $3$ subcategories: \textbf{(a) Layer Node Replacement (LN)}, replacing layers and computational nodes (e.g. replacing a \texttt{ReLU} node to \texttt{Sigmoid}); \textbf{(b) Arithmetic Node Replacement (AN)}, replacing nodes related to batch arithmetic operations (e.g., \texttt{Add} node to \texttt{Subtract}); and (c) \textbf{Node Input/Output Modifications (NIO)} (e.g., replacing a data tensor input node to a \texttt{Dense} node). All graph-related mutations are implemented on Relay IR in Apache TVM. For each subcategory, \tool\ traverses the Relay IR Abstract Syntax Tree (AST), and either replaces function calls or injects operations. 
In addition, it analyzes model metadata such as input and output dimensions across layers, and adjusts them in order to preserve mutant validity. An indicative example (transposing input tensor of a Conv2D layer) is shown in Figure~\ref{fig:transpose_example}.

\begin{figure}[!htp]
\centering
\includegraphics[width=0.82\linewidth, height=1.4cm]{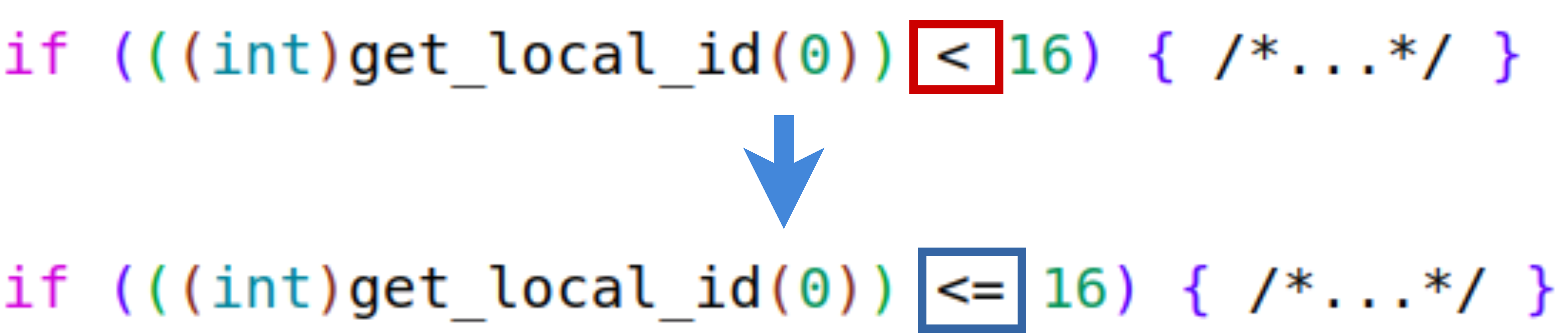}
 \caption{Mutation of a conditional operator in TIR for a fused operation in MobileNetV2.}
 \label{fig:mutation}
\end{figure}

For code-related mutations, we generate mutants based on three subcategories, focusing on operations modifying statements and constructs at a lower level: \textbf{(a) Arithmetic Types Mutations (AT)} (e.g., changing \texttt{float32} to \texttt{float16} in variable types); \textbf{(b) Kernel Variables \& Stores Mutations (SV)} (e.g., adding a constant of $0.5$ to an existing variable holding a numeric value); and \textbf{(c) Conditional Statement Operations Mutations (CSO)}, such as replacing the less-than ($<$) operator of a test expression inside a conditional statement by less-than-equal ($<=$), as shown in Figure~\ref{fig:mutation}. 
These mutations are applied at the lower-level, Tensor IR (TIR) of Apache TVM. \tool\ traverses the TIR AST, and mutates the necessary statements and expressions based on configuration.

\paragraph{Inference Analysis}
\tool\ is capable of automatically performing differential testing for all the mutants generated and deployed on different hardware acceleration devices. In particular, \tool\ analyzes and compares the execution results of varying mutants, by extracting their results and performing pairwise comparison.
For instance, consider MobileNetV2 mutated to \textit{MobileNetV2M1} and the mutant was configured to run on $3$ devices (\textit{D1}, \textit{D2}, and \textit{D3}) against a dataset defined in the configuration. 
Following execution on all the above configurations, \tool\ will identify the folders with the outputs and compare, element-wise for each dataset output, across all combinations of devices (\textit{D1\-D2}, \textit{D1\-D3}, \textit{D2\-D3}). \tool\ supports a variety of element-wise comparator strategies, such as top-1 prediction and Levenshtein distance, and Kendall's Tau~\cite{kendall}, while automatically ruling out mutant crashes.
In addition, \tool\ performs the comparisons in an efficient way, by automatically ruling out redundant comparison cases, such as where a mutant crashed, or produced unreasonable results (e.g., all inferences having the same or very similar predictions) for one or more devices - at which point \tool\ will disregard the case and proceed only with the rest of the valid runs. As a result of the analysis, \tool\ generates a number of JSON files containing metadata, such as the difference percentage across output label predictions and the dataset image output names, that presented discrepancies.

\section{Related Work}
\label{related-work}
Existing work has primarily focused on exploring DNN model correctness applying adversarial testing~\cite{deepxplore, zhang2018deeproad}. 
In terms of mutation testing, 
DeepCrime~\cite{deepcrime}, DeepMutation~\cite{deepmutation} and DeepMutation++~\cite{deepmutationplusplus} are tools that generate model mutants to assess the test data input quality for convolutional and recurrent NNs, with the first specializing in pre-training mutation strategies, and the latter two focusing on post-training mutant generation. LEMON~\cite{wang2020lemon} and CRADLE~\cite{pham2019cradle} explore mutation testing and fault localization from models sourced from different Deep Learning (DL) frameworks. 
Regarding ML compiler fuzzing, Tzer~\cite{TZER} is a coverage-guided fuzzer that applies modifications on models compiled in Apache TVM utilizing Tensor IR (TIR).
TVMFuzz~\cite{TVMFuzz} is a tool that allows generic-purpose TIR fuzzing on Apache TVM, while HirGen~\cite{HirGen} focuses on fuzzing the high-level IR of TVM (Relay IR).
NNSMith~\cite{NNSMith} focuses on the generation of valid DNN graphs that can be used to examine the compiler for bugs. 
Ren et al.~\cite{ren2023effectiverandomtestgeneration} propose a constraint solving tool for bug detection utilizing semantic specifications for AI compilers.
The primary purpose of these tools is to detect either crashing bugs, or floating-point inconsistencies without considering hardware acceleration, while our work is about detecting bugs that cause different behavior, primarily focusing in the context of hardware accelerators (inspired by a well-established taxonomy of faults~\cite{humbatova2020taxonomy}, where GPU issues are identified as a primary cause of DNN model faults). In addition, our prior research~\cite{louloudakis2022assessing, deltann} highlights unexpected performance deviations across different GPUs while using the same set of optimizations. This serves as a motivation for \tool.

\section{Evaluation}
\label{experiments}
\label{experiment-setup}

We considered $7$ widely-used image recognition models of varying sizes and architectures, used for image classification, as well as more complex activities, such as semantic segmentation and object detection:
ShuffleNet~\cite{shufflenet} ($5.46$MB, $2$M parameters), MobileNetV2~\cite{mobilenetv2} ($14$MB, $3.4$M parameters), ResNet152V2 ~\cite{resnetv2} ($230$MB, $115.6$M parameters), AlexNet~\cite{alexnet} ($233$MB, $60$M parameters), EfficientNet (Lite) ~\cite{efficientnet} ($49.5$MB, $5.3$M parameters), DenseNet121~\cite{densenet} ($31.2$MB, $8$M parameters), and InceptionV2~\cite{inception} ($43$MB, $13.6$M parameters). All models are pre-trained on ImageNet~\cite{deng2009imagenet}. We selected these models as they are mature and widely used for image recognition tasks in real-world applications, while their size and number of parameters allowed us to perform experiments in a tractable manner, allowing us deployment on hardware acceleration devices with varying capabilities (including a low-end, \texttt{Mali} GPU). We use the object detection test dataset of the ImageNet Large Scale Visual Recognition Challenge ~\cite{ILSVRC17} as our experiments base dataset, given its complexity and realistic content.
We tested our original DNN models and their respective mutants against $4$ hardware accelerators of different levels of capabilities: an Intel-based server featuring an Nvidia Tesla K40c GPU (\textit{Server \#1}) and an Nvidia Titan Xp GPU({\textit{Server \#2}}); an Nvidia AGX Xavier featuring an Nvidia Volta GPU (\textit{Xavier}); and a mobile-class Hikey 970 board featuring an Arm Mali-G72 GPU (\textit{Hikey}). 
Initially, we run the original models across devices to check for potential problems without applying any changes to them, and we detected no differences in output predictions. We then proceeded generating mutations and performing differential testing across devices.
Overall, we generated $21$ mutations covering all $6$ categories supported by \tool, as described in Section~\ref{implementation}. Our mutation set is presented in Table~\ref{tab:mutations}.

\begin{table}[!ht]
    \def\arraystretch{1.2}
    \caption{Mutations generated for evaluating \tool.}
    \centering
    \label{tab:mutations}
    \begin{tabular}{|l|l|}
    \hline
        \textbf{Cat.} & \textbf{Mutations} \\ \hline
        \textbf{LN} & ReLU $\rightarrow$ Sigmoid \\ \hline
        \textbf{AN} & Add $\rightarrow$ Subtract \\ \hline
        \multirow{2}{*}{\textbf{NIO}} & Transpose(Bias) for \textit{Conv2D}, Transpose(Input) \\ & and Exp(Input/Output) for \textit{Conv2D}, \textit{Dense}, \textit{BatchNorm} \\ \hline
        \textbf{AT} & Float32 to Float16, Int16, Int8 \\ \hline
        \multirow{2}{*}{\textbf{CSO}} & LT($<$) $\rightarrow$ LTE($<=$), GT($>$) $\rightarrow$ GTE($>=$), GT($>$) $\rightarrow$ LT($<$),\\ & $x < y \rightarrow x < (y + 0.5)$ \\ \hline
        \textbf{SV} & Var/Store  Sub(1e - 8), Mul(1 + 1e - 8) \\\hline
    \end{tabular}
    \vspace{-10pt}
\end{table}

\section{Preliminary Results}
\label{preliminary-results}
In this section, we report our observations for each of the categories of mutants, as preliminary results in order to showcase the usability of \tool. It is important noting however that our focus is primarily on the observation of model behavior differences across devices as a result of the fault present on the corresponding mutant, and not on cases where mutants resulted in identical behavior across devices. This is because essentially a mutant that is consistently problematic can identified as problematic and "killed" (either by testers manually, or from a potential testing suite) with ease, while for cases where the behavior is inconsistent across devices, the effects of a fault has a much higher chance to be overlooked. The only exception to the rule is with regard to \textit{AT} category of mutants, where the model behavior was unexpected, given the nature of the fault introduced - and which we present in more detail below.

\subsection{Graph-Related Mutants}
For graph-related mutants, we concluded with the following observations: \textbf{LN:} The replacement of \textit{ReLU} by \textit{Sigmoid} occurences in mutants led most of the cases to crash, with the exception of \textit{EfficientNet}, where the mutant achieved identical results to the original on \textit{Xavier} device, while it crashed on all other $3$ devices.
\textbf{AN:} We observed heavy model performance degradation (up to $99.8$\% on \textit{InceptionV2}) that was consistent across devices. 
\textbf{NIO:} The transposition of both data and weights of convolutional nodes (Conv2D) led to heavy model performance degradation (up to $98.4$\% on \textit{InceptionV2}), a behavior that was also consistent across devices. 
\begin{figure*}[t]
\centering
\begin{subfigure}{0.49\textwidth}
    \centering
    \advance\leftskip-0.5cm
    \includegraphics[width=1.05\linewidth, height=4.3cm]{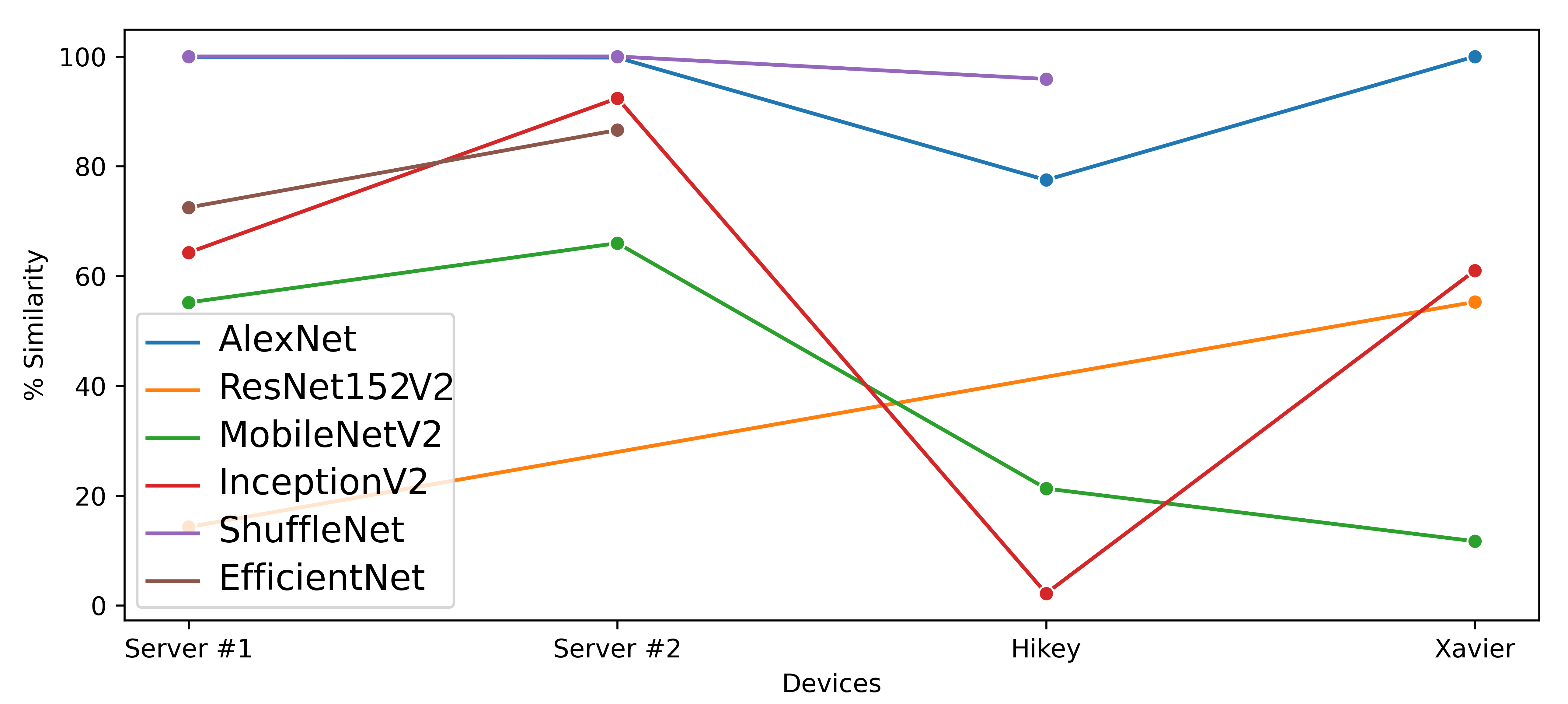}
     \caption{LT-to-LE}
     \label{fig:muta}
\end{subfigure}
\centering
\begin{subfigure}{0.49\textwidth}
    \centering
    \advance\leftskip-0.2cm
    \includegraphics[width=1.05\linewidth, height=4.3cm]{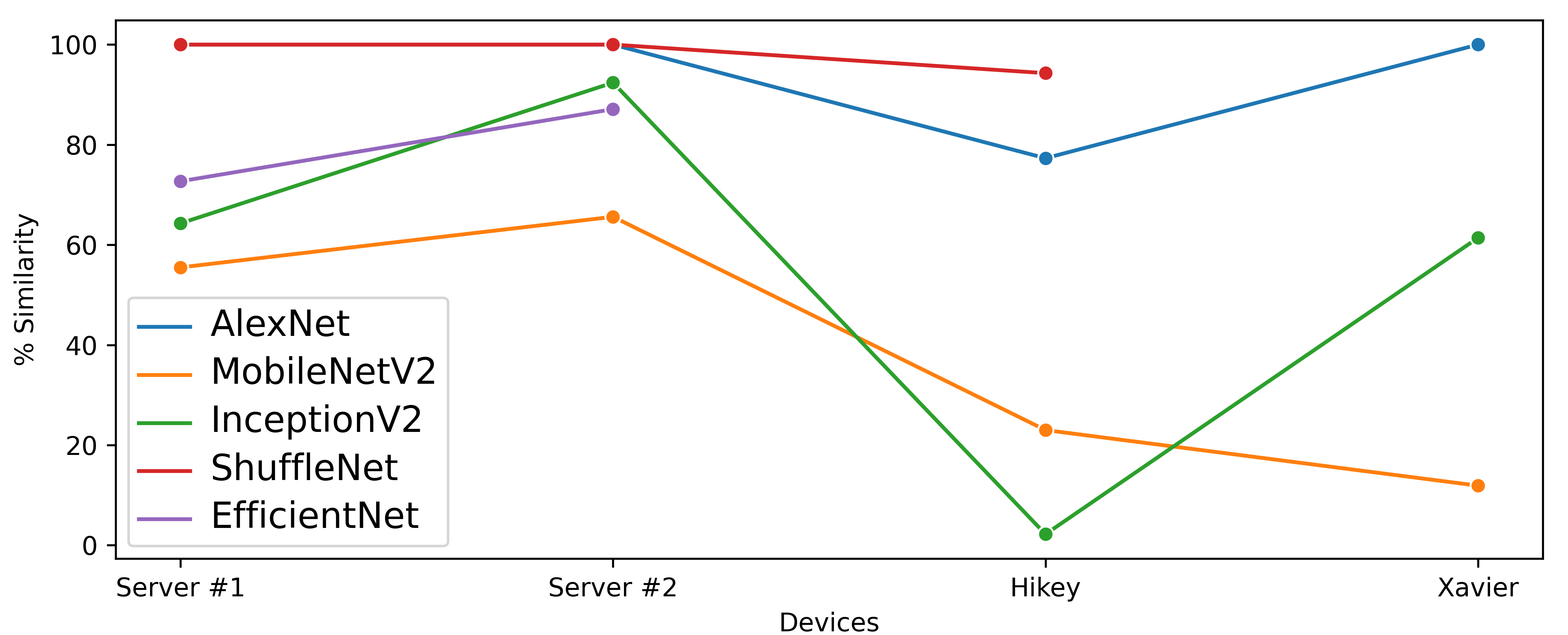}
    \caption{LT-RA}
    \label{fig:mutb}
\end{subfigure}
 \caption{Comparison of mutants against the original model across devices: (a) LT to LTE and (b) right value increased by 0.5 in conditional statements for all the DNNs under test.}
  \label{fig:muts}
  \vspace{-15pt}
\end{figure*}

\subsection{Code-Related Mutants}
For code-related mutants, the results were the following:
\textbf{AT:} The models produced predictions that instead of considering the full set of classification labels, they were limited to a very small subset of them, for all the dataset inputs, misclassifying the vast majority of them.
For example, for \textit{EfficientNet}, the mutant of changing arithmetic types from \textit{Float32} to \textit{Int16}, distinct images containing a goldfish and an owl, are both mistakenly identified as ``paddle wheel", essentially resulting in a not crashed, but practically unusable model.
Although this result was consistent across devices, we considered it interesting, as many optimizations (e.g., Fast-math) compromise model accuracy over computation speed and size. Furthermore, a change in arithmetic types (e.g., \textit{Float32} to \textit{Int16}, was expected to produce an effect related to only partial accuracy drop, due to loss of precision, a considerably different behavior in comparison to the one observed in our experiments with \textit{AT} mutants, as aforementioned.
\textbf{SV:} The mutations seemed to have no effect to the model correctness. However, our constant mutations in store values were limited in the models. We consider applying more drastic changes in future work.
\textbf{CSO:} we observed different behavior across hardware acceleration devices in cases of (1) operator modifications and (2) changes in conditional thresholds. For (1), and although most of the conditional operator mutants led to similar results across devices, the mutation changing the less-than (LT) to less-than-equal (LTE) operator in conditional statements (LT-to-LE) produced up to $90.3$\% discrepancy, observed in InceptionV2 across~\textit{Server \#2} and \textit{Hikey}.
For (2), an increase of the right operand value of a conditional statement using LT by a constant (LT-RA) presented different behavior across devices, with once again \textit{InceptionV2} presenting a difference of up to $90.2$\% across the same devices with (LT-to-LE). Mutations related to the greater-than (GT) operator had no effects to the models across devices, with the exception of \textit{EfficientNet}, which crashed on \textit{Hikey}.
The results of both mutants can be observed in Figure~\ref{fig:muts}, following a similar trend, as both of them essentially modify the lower bound of a LT condition. For the rest of the mutants related to conditional operators, no significant results were observed.

\subsection{Results Overview}
Overall, we conclude to the following observations: (1) Differences in kernel code related to conditional operations can affect output label correctness across devices, in varying degrees.
(2) Heavy modifications in model types associated with precision change in kernel code (\textit{AT} mutants) can lead to severe model performance degradation, resulting the models giving similar output classifications for completely unrelated inputs, resulting in practically unusable models. (3) For a wide variety of graph and kernel-related faults introduced, the behavior is consistent across devices (including cases where mutants crashed across devices). Observation \#2 also falls into this category, however it results in an unexpected, peculiar behavior across the models. Furthermore, we consider it as a separate, noteworthy observation. Both \#1 and \#2 observations can be attributed to the differences in implementation across programming models and GPU architectures~\cite{comparative-study-gpu}. However, exploring and identifying precisely the aspects of the observed behavior across each specific device, is out of the scope of this work.

\section{Tool Usability}
\tool\ can by used both by developers and researchers to test the model architecture and structure robustness, by simulating specific erroneous scenarios that can be introduced on model deployment or  optimization. Using \tool\ requires only basic knowledge on DNN deployment from its users.

\tool\ operates through an automated pipeline, using a JSON-based configuration, containing parameters related to all the DNN models under test, the mutations to be generated, as well as the hardware acceleration devices the mutants will be deployed on, and the datasets utilized for differential testing. \tool\ is extensible to apply more mutations through configuration, while supporting generic models deployment and inference on different devices. Model developers can configure the system to modify specific structural components of the model, as well as direct the system to apply mutations in targeted model elements.

In addition, since \tool\ is built on top of Apache TVM, a generic purpose ML compiler, it theoretically supports mutation testing on DNN models outside the image recognition domain, such as text classification. \tool\ is able to automatically fetch a plethora of pre-trained models from well-established libraries, such as PyTorch~\cite{pytorch} and Keras~\cite{chollet2015keras}, while also supports file-based custom model representations, such as ONNX~\cite{onnxsite}.

In summary, \tool\ provides a highly configurable and extensible mechanism to apply differential testing across hardware acceleration devices.
Furthermore, we propose the following use cases for \tool.
\textbf{(1) Mutants Generation to Examine Test Suite Effectiveness:} \tool\ can be utilized in order to generate mutants, with the purpose of examining the effectiveness of test suites to effectively detect faults and contribute towards model robustness across different hardware acceleration devices.
Following the traditional mutation testing paradigm, \tool\ can generate mutants that examine DNN model behavior in the context of hardware acceleration. \textbf{(2) Simulation of Errors to Examine Model Behavior:} \tool\ can be used to examine how DNNs behave in the presence of undesired faults. For instance, imagine that a conditional statement error in the kernel of a model is introduced by a developer, and this model is compiled and deployed as part of two automated driving systems that use the same DNN model, but different sets of GPUs for calculations. Using \tool, DNN model developers can simulate the effects of the error across different hardware acceleration devices, collect metrics and observe the severity of impact that such errors have. This can help them develop fail-safe mechanisms that prevent catastrophic scenarios.

\section{Threats To Validity}
There are four main factors we consider as threats to validity: (1) The mutations applied in the experiments are applied extensively in the models, for every occurrence of an operation. While this is done in order to maximize the potential of problems and effectively highlight it in this paper, the tool supports the application on mutants on specific parts of the model, subject to developer testing needs.  (2) We focused on Convolutional Neural Networks (CNNs) utilized for image recognition in our experiments. We selected to do so, given that CNNs are widely utilized in a spectrum of real-life applications (e.g., autonomous vehicles~\cite{selfdrivingcarsoverview}), while their reasonable size and number of properties allow us to conduct experiments in a tractable manner. However, our methodology can be extended to other DNN types, given that (I) mutations corresponding to the properties of such models and (II) a suitable output comparator is defined, for differential testing purposes.
(3) All models are trained to one specific dataset (ImageNet): Although the usage of additional datasets would indeed contribute to the comprehension of our experiments, this is a preliminary evaluation in order to demonstrate the capabilities of our tool. \tool\ supports the execution of different models and datasets based on the configuration defined. In addition, ImageNet is a comprehensive dataset with over 14 million images annotated and extensively used for research and development purposes and therefore a good candidate for evaluation.
(4) Randomness in the results: in order to avoid the possibility of randomness introduced in the observed results, we (1) performed multiple executions of the models on a smaller dataset consisting of $10$ images that examine different classification predictions, and verified the consistency of the outputs, as a small-scale evaluation of our experiments in total, and (2) randomly sampled a subset of our experiment runs and repeated it $3$ times. We examined the results for differences across the runs, and we did not observe any deviations attributed to randomness. As a result, we are confident that our results are consistent and not random.

\section{Discussion}
From our findings we can infer the following considerations.

\textbf{(1) Differences in Conditional Operators:} Conditional operators in kernels are quite commonly used to indicate thread computations grouping. A quite common problem in conventional software kernels is branch divergence~\cite{branchdivergence}, which mostly affects parallelization efficiency. However, our observations highlight that problematic conditional statements can also affect the predictions of the DNN model across different devices in a drastic manner and to a varying extent.
For example, Fig.~\ref{fig:muta} demonstrates that an LT-to-LE mutant generated for \textit{AlexNet} run on a device like \textit{Server \#1}, achieves results of $\approx100$\% accuracy compared to the original. However, the accuracy of the same mutant deployed on \textit{Hikey} drops to $77.5$\%. Similarly, for \textit{ShuffleNet}, the same mutant gives $100$\% accuracy compared to the original on \textit{Server \#1}, but $95.85$\% on \textit{Hikey}. If a test suite used to check model correctness allowed an accuracy deviation threshold of just $1$\%, and the models were not tested on low-end devices such as \textit{Hikey}, both mutants would probably survive. As a conclusion, we recommend that model developers extensively test for such defects across a plethora of devices, while enforcing strict thresholds in terms of acceptable accuracy deviations across devices.

\textbf{(2) Effects of Numeric Type Precision:} Usually, precision errors are expected to affect model correctness. For instance, quantization~\cite{Jacob2017QuantizationAT, liu2021posttrainingquantizationvisiontransformer} aims to improve the computation times of a model at an acceptable cost of model accuracy. However, we observed that such modifications, when applied in large scale, can affect the model so drastically that they can turn it essentially unusable. As a result, we suggest to the DNN model developers to be cautious when aiming to deploy models for hardware acceleration, as modifications that are related to changes in numeric precision for purposes such as optimization (e.g., quantization~\cite{quantization}) can potentially have a much more adverse effect to model accuracy than the initially anticipated. Furthermore, model developers should test the models to this direction across hardware acceleration devices.

\section{Conclusion}
We presented \tool, a tool that enables mutation testing of DNN image recognition models. The tool supports mutations related to model graph, tensor management, arithmetic types and kernel memory management. We demonstrated the potential of our tool by running preliminary experiments of $7$ perception AI models, generating $21$ mutations for each and executing in $4$ hardware acceleration devices of varying capabilities. We observed up to $90.3$\%  output label deviations in mutants related to conditional operators. We also observed a number of unexpected crashes in relation to numeric types.

\section{Availability}
\label{availability}

The source code of \tool\ can be found at~\underline{\small \url{
https://www.github.com/luludak/MutateNN}}.

\section*{Funding Acknowledgement}

The research has been funded by the Royal Society Industry Fellowship "AutoTest: Testing Autonomous Vehicle Perception Safety on Hardware Accelerators" and Huawei Edinburgh Joint Lab Project "RobustCheck: Testing Robustness of Compiler Optimisations and Deep Learning Frameworks".

\makeatletter
\renewcommand{\em}{}
\makeatother

\bibliographystyle{abbrv}
\bibliography{00-main.bib}

\end{document}